# Network Analysis for Explanation

**Hiroshi Kuwajima**
DENSO CORPORATION
`hiroshi_kuwajima@denso.co.jp`

**Masayuki Tanaka**
National Institute of Advanced Industrial Science and Technology
`masayuki.tanaka@aist.go.jp`

## Abstract

Safety critical systems strongly require the quality aspects of artificial intelligence including explainability. In this paper, we analyzed a trained network to extract features which mainly contribute the inference. Based on the analysis, we developed a simple solution to generate explanations of the inference processes.

## 1 Introduction

Recently artificial intelligence including machine learning and deep learning achieves unprecedented performance on an array of tasks [11, 21, 26, 27, 34]. Performance is not a sufficient condition for safety critical systems such as automated driving [1, 20, 28]. Comprehensive quality assurance is now one of the most important issues for artificial intelligence used in safety critical systems [17], however, there is no basic consensus on the set of quality indicators other than performance. Explainability is one of the keys for the quality of artificial intelligence. Artificial intelligence is often recognized as black-box with its increasing complexity, whereas there is strong social needs for explainability or interpretability on safety critical systems [3, 19, 29]. DARPA started Explainable AI program in 2017 [13], and a corresponding workshop was held. On the other hand, EU will let General Data Protection Regulation (GDPR) effect, and it will legally mandate a 'right to explanation' for automated decision making [9]. European Civil Law Rules in Robotics also includes transparency in its ethical principles [6].

The main contribution of this paper is that we propose an approach for explaining inference processes in visual recognition tasks [16] with a simple prototype solution on a convolutional neural network [10, 18].

## 2 Related Works

There are related works widely varying from visualization [2, 4, 8, 12, 23, 24, 30] to verbalization [14, 32, 33] and explanation [15, 25].

Visualization is recently a popular technology area in deep learning [12, 23]. Earlier studies are basically identify attention (focus) areas of input data in numerical ways similar to heat maps [4, 24, 30]. It is informative where it explains which areas the model is looking into [5], but it does not explain the inference process. There is another type of works focusing on visual attributes [8], not the input data. Visual attributes for each node were analyzed, and it was revealed that low level attributes such as black, brown, and furry are related with neural network nodes. Network Dissection interprets receptive fields as with visual attributes of neural networks, and quantified the interpretability by using the number of visual semantic concepts learned at each hidden unit [2].

Caption generation is a verbalization method, which describes an image [32, 33]. It describes the input image by a sentence which consists of the objects appearing in the image. It provides *what* objects are in the image, which are not the explanations of the inference processes *why* the objects are detected. There is a work clearly aiming at explanations: visual explanations [14]. It generates the post-hoc explanations of the inference results [22], where plausible explanations are generated by an explaining model independent from the original inference model. It generates human readable explanation, however using a mechanism which does not reflect the original inference process.

Pointing and Justification-based Explanation is one of the latest explanation methods to provide both true and plausible explanations [25]. It provides attention areas of input data space as introspective explanations (true explanation) and visual explanations (plausible explanation) at the same time. It provides explanations of the input space, but does not provide them for the inference processes. Hu et al. proposed generated network structures easy to understand and explain, however, it cannot apply to pre-trained models with specific structures [15].

## 3 Explaining Inference Process

As seen in related works, there is no research to provide explanations of the inference processes so far. We propose an approach of explaining inference processes: the explanations of the inference processes are provided along with the inference results; explanations are generated for pre-trained models without retraining; explanations are prepared using the training data set or any other data sets other than the testing data set; both inference results and corresponding explanations are evaluated in the testing data set. In this section, we propose a simple algorithm to explain inference processes. It does not require retraining, and is based on the statistical natures of any one of intermediate features in a pre-trained model for a training data set. We carry out both training time and testing time feature analysis to obtain activated features, class frequent features, and inference explaining features, as described in 3.1. Then, explanations of inference processes are generated based on these newly introduced features and human annotated visual attributes, as described in 3.2.

### 3.1 Feature Analysis

We introduce three principles for explaining inference processes: 1. the highly activated features affect the inference results; 2. the features closely associated with the classes of the inference results are the basis of the inference results, 3. the overlaps between these two types of features explain the inference processes. Based on the principles, we propose three concepts called an activated feature for each inference, a class frequent feature for each class, and an inference explainable feature for each inference, as depicted in Fig. 1a, 1b and 1c respectively. We illustrate them with a specific example of CaffeNet and its `conv5` feature, where CaffeNet [7] is a modified AlexNet [18].

To explain inference processes, we focus on the activations of an intermediate feature called `conv5` which is the final convolved feature in CaffeNet. It is reported that humans are better at recognizing and agreeing upon high level visual concepts such as objects and parts in `conv5` in AlexNet [2]. Let $x$ and $y$ are the input and the output of CaffeNet. Specifically $x_i^{\text{train}}$, $y_i^{\text{train}}$ and $x^{\text{test}}$ are those of the training data and the testing data. **Activated feature** $a$ in Fig. 1a is the binarized `conv5` feature. We first apply a global max pooling to `conv5` of size $13 \times 13 \times 256$, to obtain 256-dimensional feature vector, $bmz$. Then we compute the mean-normalized feature vector $\hat{z}$, as each element of $z$ has varying dynamic range and normalization makes elements comparable each other. Thresholding $\hat{z}$ at $\gamma$ gives a binarized feature vector $a$ corresponding to $x$. **Class frequent feature** $q$ in Fig. 1b is the set of binary vectors indicating the frequently activated feature for each class. Each class has a different frequent activation pattern. Fig. 1b shows how to compute the class frequent feature for an example class: dog. The training data $x_i^{(\text{train})}$ of the dog class is binarized into $a_i^{(\text{train})}$, and their summation over $i$ counts how many times each element of the feature is activated for the dog class in the training data. After summation , we select the top-$k$ frequent elements for the class frequent feature, where $k = 3$ in the case of Fig. 1b. Class frequent features are computed for each class at the trianing time, and stored in a lookup table to be used in the testing time, like $q(\text{dog}) = [1, 0, 1, 0, 1]$, $q(\text{cat}) = [1, 1, 0, 0, 1]$ and $q(\text{bird}) = [1, 0, 0, 1, 1]$. **Inference explainable feature** $e^{\text{test}}$ in Fig. 1c is the overlap between the activated feature $a^{\text{test}}$ and the class frequent feature $q^{\text{test}}$ for a single $x^{\text{test}}$, where $\otimes$ denotes element-wise product. The dotted box in Fig. 1c is the conventional inference without explanation. $a^{\text{test}}$ is computed based on $x^{\text{test}}$, whereas the class frequent feature $q^{\text{test}}$ is, as the ground truth is unknown in the testing time, lookup by the inference result $y^{\text{test}}$ by CaffeNet. We assume the inference explainable features are the basis for the inference processes.

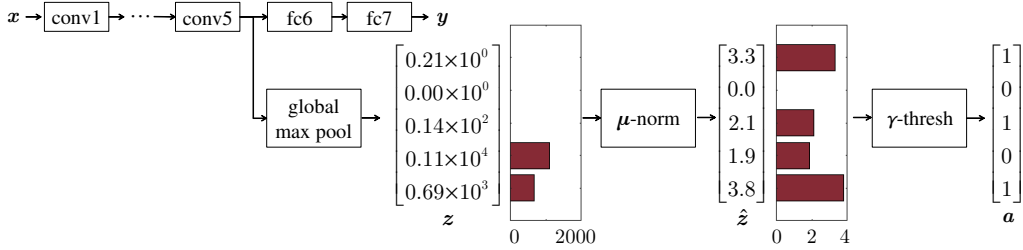

(a) Activated Feature

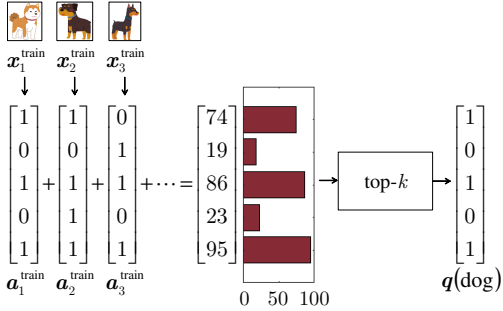

(b) Class Frequent Feature

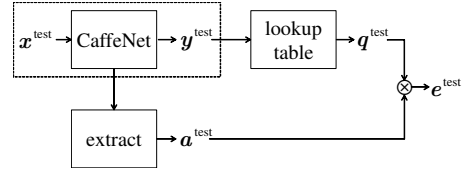

(c) Inference Explainable Feature

Figure 1: Feature Analysis Methods

### 3.2 Generation of Inference Explanations

First, we extract inference explainable features for each inference by the procedure shown in 3.1. Then, we show the inference result and explanations associated to the extracted inference explainable features. To generate human readable explanations, we annotate visual attributes for each element of the feature used for explaining inference processes, by looking at the activated features. Visual attributes and elements of the feature are generally in a many-to-many relationship, i.e., multiple elements of the feature can represent a single visual concept, and vice versa. Multiple visual attributes are annotated to each element of the feature in OR condition, because either of these visual attributes may appear for each element of the feature. We generate explanations of each active element of inference explainable features, such as "It has tiger patterns, two-tone black/brown or furs." As inference explainable features generally have varying numbers of multiple active elements, due to the human interpretability, we decided to show the elements with at most top-$\ell$ mean-normalized activations. The numbers of active elements in inference explainable features can be less than $\ell$. Now we generate explanations of the inference processes by listing the explanations of each activated elements of the inference explainable features, where $\ell = 3$, such as "This is a cat, because 1) it has tiger patterns, two-tone black/brown or furs; 2) it has animal hands or brown color; 3) it has furry surfaces, furs or animal ears."

## 4 Experiments

We try to generate explanations of the inference processes of the publicly available CaffeNet with the weighs pre-trained on ImageNet. Although ImageNet has approximately 1300 training images per class, for simplicity, we selected 100 examples for each class, with top-100 softmax probability on the ground truth classes. Selected 100 training images per class are used for computing the `conv5` feature mean, class frequent features and annotating visual attributes by human. On the other hand, we reduced the 1000 object categories of ImageNet to 32, because it is difficult for human to distinguish 1000 categories and understand the corresponding explanation precisely. The 32 classes are a subset of ImageNet 1000 classes, which are programmatically selected according to the WordNet hierarchy, such that each new class has approximately the same number of WordNet synsets. Generated sample explanations of inference processes are shown in Fig. 2.

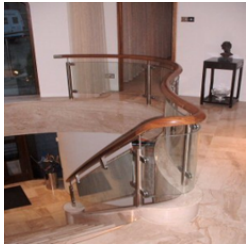
(a) This is bannister (obstruction) because, 1) it has fine lattice patterns, thin rods, fine horizontal/vertical lines or rectangle lattice patterns; 2) it has fine lines; 3) it has square or square windows.

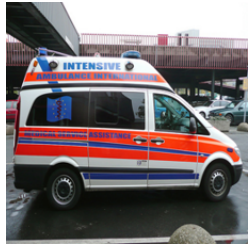
(b) This is ambulance (motor vehicle) because, 1) it has rubber tires or heads of birds; 2) it has accumulated fine boxes/circles, large characters or two-tone red/white; 3) it has faces of black dogs or black square windows.

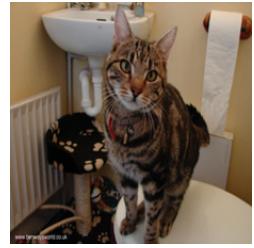
(c) This is tabby (feline) because, 1) it has leopard patterns, faces of small animals, furs or two-tone brown/white; 2) it has animals, furs or two-tone black/brown; 3) it has furs or two-tone black/gray.

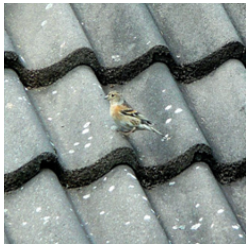
(d) This is thunder snake (snake) because, 1) it has squiggle; 2) it has squiggle or standing ears of dogs and cats.

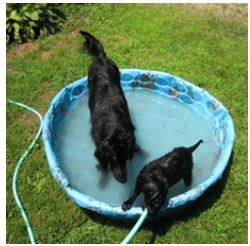
(e) This is drake (aquatic bird) because, 1) it has accumulated fine boxes/circles, green or blue; 2) It has rounded or green; 3) It has colorful colors or blue.

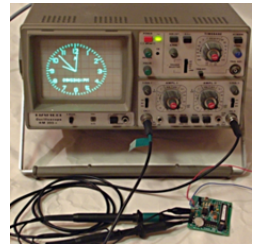
(f) This is cassette player (electronic equipment) because, 1) it has rounded; 2) it has accumulated fine boxes/circles or rubber tires; 3) it has rubber tires or square windows.

Figure 2: Explanations of Inference Processes

Figure 2a, 2b and 2c shows explanations of successful inference results. At least one visual attributes in OR conditions appear in the images, and the explanations are convincing to human. We observe three types of visual attributes 1. shape (fine lattice patterns, accumulated fine boxes/circles, leopard patterns), 2. color (two-tone red/white) and 3. concrete object (black square windows, faces of small animals). Convincing explanations of unsuccessful inference results are selected in Fig. 2d, 2e and 2f, where the ground truth labels are brambling (passerine), flat-coated retriever (sporting dog) and analog clock (measuring instrument). 2d has a bird on the roof tiles, however, the edges between tiles squiggle and look like snakes. We can see the color visual attributes such as green and blue mislead the inference in 2e. The unsuccessful inference result is cassette player for 2f, however the explanation of the inference process is understandable for what actually appears in the image: an oscilloscope. This example shows that the explaining inference processes can reveal the limitation of CaffeNet or mistakes in labeling.

## 5 Conclusion

We proposed a simple approach of explaining inference processes, and developed an explaining system. The system was qualitatively evaluated. Our developed system is one of the most simplest solution for explaining inference processes. We expect research communities to tackle the problem of explaining inference processes and improve the methods, because it is an important enabling technology for artificial intelligence to be deployed in safety critical systems.